\pdfoutput=1
\documentclass[11pt]{article}

\usepackage[]{acl} 

\usepackage{times}
\usepackage{latexsym}

\usepackage[T1]{fontenc}
\usepackage[utf8]{inputenc}

\usepackage{microtype}

\usepackage{amsmath}
\usepackage{url}
\usepackage{graphicx}
\usepackage{amsfonts}
\usepackage{CJK}
\usepackage{bm}
\usepackage{mathrsfs}
\usepackage{float}
\usepackage{xcolor,colortbl}
\usepackage{subfigure}
\usepackage{booktabs,multirow}
\usepackage{bigstrut,bigdelim}
\usepackage{paralist}
\usepackage{bbm}
\usepackage{helvet}
\usepackage{courier}
\usepackage{makecell}
\usepackage{nicefrac}
\usepackage{epsfig}
\usepackage{diagbox}
\usepackage{framed}
\usepackage{empheq}
\usepackage{array}
\usepackage{enumitem}
\usepackage{mdwlist}
\usepackage{xspace,mfirstuc,tabulary}
\usepackage{tcolorbox}
\usepackage{algorithm}
\usepackage{placeins}
\usepackage{algpseudocode}
\DeclareMathOperator*{\softmax}{softmax}

\newcommand{\MODELNAME}{\textsc{TegTok}}
\title{\textsc{TegTok}: Augmenting Text Generation via Task-specific \\and Open-world Knowledge}

\author{Chao-Hong Tan$^1$\thanks{\hspace{1.5mm}Work done during the internship at Microsoft.}, Jia-Chen Gu$^1$\footnotemark[1], Chongyang Tao$^2$, Zhen-Hua Ling$^1$\thanks{\hspace{1.5mm}Corresponding author.}, \\
{\bf Can Xu$^2$, Huang Hu$^2$, Xiubo Geng$^2$, Daxin Jiang$^2$\footnotemark[2]} \\
  $^1$National Engineering Research Center for Speech and Language Information Processing, \\
      University of Science and Technology of China, Hefei, China \\
  $^2$Microsoft, Beijing, China \\
{\tt \{chtan,gujc\}@mail.ustc.edu.cn}, {\tt zhling@ustc.edu.cn}, \\ {\tt \{chotao,caxu,huahu,xigeng,djiang\}@microsoft.com}
}

\begin{document}
\maketitle
\begin{abstract}
  Generating natural and informative texts has been a long-standing problem in NLP. 
%   Existing studies have achieved impressive results through fine-tuning the large-scale pre-trained language models (PLMs) on downstream tasks. 
%   However, {these models  ability to access and manipulate knowledge is still limited}. 
%   Although much effort has been dedicated into incorporating PLMs with various open-world knowledge, such as knowledge graphs or wiki pages.
  Much effort has been dedicated into incorporating pre-trained language models (PLMs) with various open-world knowledge, such as knowledge graphs or wiki pages. 
%   {open-word knowledge is sparse and exists in PLMs, it's not easy to acquire the knowledge from the specific task}
%  However, on many downstream tasks, this type of world knowledge is not always necessary
However, {their ability to access and manipulate the task-specific knowledge is still limited} on downstream tasks, as this type of knowledge is usually not well covered in PLMs and is hard to acquire.
%   \cy{especially when the training data is limited or the distribution of downstream data is  different from the original model}.
  To address the problem, we propose augmenting \textsc{TE}xt \textsc{G}eneration via \textsc{T}ask-specific and \textsc{O}pen-world \textsc{K}nowledge (\textsc{TegTok}) in a unified framework. 
  Our model selects knowledge entries from two types of knowledge sources through dense retrieval and then injects them into the input encoding and output decoding stages respectively on the basis of PLMs. 
  With the help of these two types of knowledge, our model can learn what and how to generate. 
  Experiments on two text generation tasks of dialogue generation and question generation, and on two datasets show that our method achieves better performance than various baseline models.

%  Generating natural and informative texts has always been a challenge for building text generation systems. Recently studies have investigated achieving this goal by incorporating generative models with a variety of world knowledge, such as knowledge graphs or documents from Wikipedia.
%   However, this type of world knowledge is not always necessary since the relevant knowledge is sparse in many cases. 
%   Open-domain texts, on the other hand, are often grounded by more than one kind of knowledge perception. 
%   In addition to world knowledge, task-specific knowledge also acts as an important information source.
%   Intuitively, the related task-specific examples from a pre-built index can also provide exemplary information for neural generative models when generating according to a given input sequence. 
%   To simulate this motivation, we propose augmenting \textsc{TE}xt \textsc{G}eneration via \textsc{T}ask-specific and \textsc{O}pen-world \textsc{K}nowledge (\textsc{TegTok}). %(documents from the Wikipedia) and  knowledge (responses from a pre-built index). 
%   Our model selects knowledge entries from two types of knowledge sources through dense retrieval and then injects them into the input encoding and output decoding stages respectively. 
%   With the help of these two types of knowledge, our model can learn what and how to generate in a unified framework. 
%   Experiments on two text generation tasks of dialogue generation and question generation, and on three datasets show that our method achieves better performance than various baseline models. 
\end{abstract}

\section{Introduction}

  Enabling natural models to generate natural and informative sequences is a challenging yet intriguing problem of artificial intelligence and has attracted increasing attention due to its promising potentials and alluring commercial values~\cite{DBLP:journals/corr/BahdanauCB14,DBLP:conf/acl/DuSC17,DBLP:conf/ccwc/KepuskaB18,DBLP:conf/ucami/Berdasco0D0G19,DBLP:journals/coling/ZhouGLS20,DBLP:journals/corr/abs-2102-01672}. 
%   A basic definition of the text generation task is to generate an expected output sequence from a given input sequence, called sequence-to-sequence (Seq2Seq)~\cite{DBLP:conf/nips/SutskeverVL14}.
  Thanks to the achievements on neural sequence modeling and pre-training technologies, current generative models are able to generate nature and fluency target sequences using either encoder-decoder architectures~\cite{DBLP:conf/nips/SutskeverVL14,DBLP:journals/corr/BahdanauCB14,DBLP:conf/nips/VaswaniSPUJGKP17} or language models~\cite{radford2019language,DBLP:conf/nips/BrownMRSKDNSSAA20,DBLP:conf/acl/LewisLGGMLSZ20}
%   endowing machines the ability to return natural and informative sequences. %~\cite{DBLP:conf/acl/ShangLL15,DBLP:conf/aaai/SerbanSBCP16,DBLP:conf/aaai/SerbanSLCPCB17}
  Despite these methods being the state-of-the-art frameworks for NLG, 
  they are often provided limited knowledge to generate the desired output. 
%   Meanwhile, some existing methods such as the attention~\cite{DBLP:journals/corr/BahdanauCB14} and copy mechanism~\cite{DBLP:conf/acl/GuLLL16} generally suffer from an inability to well retain and recall knowledge.
  Thus, the performance of text generation is still far from satisfaction in many real-world scenarios~\cite{DBLP:journals/corr/abs-2010-04389}.
%   \cy{please refer to Retrieval-Augmented Generation for Knowledge-Intensive NLP Tasks \& A Survey of Knowledge-Enhanced Text Generation. ``generic and boring'' are not proper for traditional text generation. The following paragraph also need to be more general.}
  
    % generative models with a variety of world knowledge, such as knowledge graphs or documents from Wikipedia.
  Recently, much effort has been dedicated into incorporating traditional generative models or pre-trained language models (PLMs) with a variety of open-world knowledge, such as structural knowledge bases (e.g., ConceptNet)~\cite{DBLP:conf/lrec/SpeerH12,DBLP:conf/aaai/SpeerCH17} or unstructured documents (e.g., documents from Wikipedia)~\cite{DBLP:conf/emnlp/ZhouPB18,DBLP:conf/iclr/DinanRSFAW19}.
  By providing the supplementary knowledge of an entity mentioned within or the background knowledge of a source text, it can help to better understand the input text and its surrounding context, and to ameliorate the informativeness of the generated text.

  Although the open-world knowledge brings improvement to the generation process in most cases, its effect is still limited to the cases involving fewer entities or abstract semantics.
  On the other hand, the process of generating text by humans is often grounded by more than one single type of knowledge perception. 
  In addition to world knowledge, the task-specific knowledge also acts as an important information source, and is usually not well covered in PLMs and is hard to acquire through fine-tuning.
  For example, in dialogue systems, what people have said or responded before can be reused as an important knowledge source, where these utterances talked before can be retained as the task-related knowledge in the mind of an interlocutor;
  for question generation, what part of a document makes people curious most and then ask specific questions, can often get enlightened by the existing questions raised from their corresponding passages.
%   for summarization, passages sharing the same task-specific topic might contain the similar keywords which can be used to enrich and modify a summary to ensure grammatical correctness and fluency.
  Intuitively, these related task-specific examples can bring additional information associated with the given source messages and provide exemplary information for neural generative models, but this useful information source is neglected in previous studies. 
%   Most of the existing studies focus on integrating only one single type of knowledge, while an important but under-explored field in text generation is to leverage and combine different types of knowledge together.

  % our proposal 
  On account of the above issues, we propose augmenting \textsc{TE}xt \textsc{G}eneration via \textsc{T}ask-specific and \textsc{O}pen-world \textsc{K}nowledge (\textsc{TegTok}). % \textbf{A}ugmenting \textbf{T}exts via \textbf{T}ask-specific \textbf{A}nd \textbf{C}ommon-world \textbf{K}nowledge (\textbf{ATTACK}). 
  Specifically, the world knowledge is assumed to be \emph{unstructured Wikipedia documents} that provide supplementary information of an entity mentioned within or background knowledge of an input sequence.
%   is domain relevant and
%   that consistent with the form of the task definition 
  The task-specific knowledge is a pre-built index that is domain-relevant and acts as an exemplary information source for guiding text generation. It can be flexibly adjusted according to different tasks or domains, e.g., context-response pairs in dialogue generation or passage-question pairs in question generation.
%   is specified as \emph{a pre-built retrieval index} that acts as an exemplary information source for guiding text generation.
  Inspired by the success of dense retrieval methods for the task of open-domain question answering~\cite{DBLP:conf/acl/LeeCT19,DBLP:conf/icml/GuuLTPC20,DBLP:conf/emnlp/KarpukhinOMLWEC20}, we use pre-trained encoders to convert input texts and knowledge entries into dense representation vectors and employ fast maximum inner-product search (MIPS) \cite{DBLP:conf/nips/Shrivastava014} to complete the retrieval, so as to ensure effectiveness and efficiency of knowledge selection.
  Finally, these two types of knowledge are injected into source text encoding and target text decoding stages respectively. 
  By this means, our model can learn how and what to generate in a unified framework with the help of two types of knowledge.

  To measure the effectiveness of our proposed framework, we evaluate it on the tasks of dialogue generation and question generation, which are both important research issues of text generation. 
%   \MODELNAME{} is evaluated on the Reddit dataset \cite{DBLP:conf/ijcai/ZhouYHZXZ18} in the task of dialogue generation, and on the SQuAD dataset \cite{DBLP:conf/acl/DuSC17} in the task of question generation. % Wizard of Wikipedia dataset \cite{DBLP:conf/iclr/DinanRSFAW19}
  Experimental results show that our proposed method outperforms the GPT-2 \cite{radford2019language} and BART \cite{DBLP:conf/acl/LewisLGGMLSZ20} baseline models, and can generate more informative texts including entities that do not appear in the input texts.
%   \emph{To facilitate the reproduction of our results, we will publish all source code later.}
  
  In summary, our contributions in this paper are three-fold: 
  (1) A proposal of a general and unified text generation framework named \MODELNAME{} that incorporates both task-specific and world knowledge through dense retrieval. 
  (2) The proposed framework is verified on two text generation tasks. %and on three datasets.
%   (3) Experimental results verify the effectiveness of our proposed method on both automated and human evaluation metrics.

\section{Related Work}
  
  \vspace{-1mm}
  \paragraph{Knowledge-enhanced Text Generation.}
%   This task aims at incorporating the knowledge to enhance the generation of the output text through leveraging the dependencies among the input text, knowledge, and output text.
%   In NLG systems, knowledge is an awareness and understanding of the input text and its surrounding context.
  As knowledge can help to understand the input text and its surrounding context, many previous studies explored the leverage of knowledge bases~\cite{DBLP:conf/lrec/SpeerH12,DBLP:conf/aaai/SpeerCH17,koncel-kedziorski-etal-2019-text,liu2021kg} or unstructured texts~\cite{DBLP:conf/acl/KielaWZDUS18,DBLP:conf/emnlp/ZhouPB18,DBLP:conf/iclr/DinanRSFAW19,lewis2020retrieval} for the text generation task, and they have demonstrated promising performance on generating informative and coherent texts. To incorporate unstructured knowledge from the web, retrieval-augmented text generation~\cite{lewis2020retrieval} has been widely explored. Besides, researchers also introduced the paradigm of retrieve-and-edit~\cite{DBLP:conf/nips/HashimotoGOL18,wu2019response,ren2020retrieve} or exemplar-based decoding~\cite{DBLP:conf/naacl/PengPFD019,gupta2020controlling} to enhance the generation processes with similar input-output pairs come from the specific task. More related works about knowledge-enhanced text generation can be referred to \citet{yu2020survey}.
%   In this paper, we focus on evaluating the proposed \MODELNAME{} model on the tasks of \emph{dialogue generation} and \emph{question generation}.

  \vspace{-1mm}
  \paragraph{Dialogue Generation.}
%   The existing neural dialogue models can be generally categorized into retrieval-based and generation-based methods.
%   The retrieval-based methods learn a matching model for a pair of a conversational context and a response candidate~\cite{DBLP:conf/acl/WuWXZL17,DBLP:conf/acl/WuLCZDYZL18,DBLP:conf/wsdm/TaoWXHZY19,DBLP:conf/cikm/GuLL19,DBLP:conf/cikm/GuLLLSWZ20,DBLP:conf/aaai/XuTJZ0021}.
  
  The generation-based dialogue models synthesize a response with a NLG model by maximizing its generation probability given the previous conversation context.
%   This approach enables the incorporation of rich context when mapping between consecutive dialogue turns \cite{DBLP:conf/acl/ShangLL15,DBLP:conf/aaai/SerbanSBCP16,DBLP:conf/emnlp/LiMSJRJ17}.
  The pioneer researchers formulated the dialogue generation task as a sequence-to-sequence translation problem~\cite{DBLP:conf/acl/ShangLL15,DBLP:conf/naacl/SordoniGABJMNGD15,DBLP:journals/corr/VinyalsL15,DBLP:conf/aaai/SerbanSBCP16,DBLP:conf/aaai/SerbanSLCPCB17} where encoder is designed for dialogue context modeling, and decoder is constructed to conduct the target response prediction. 
  Expanded from the general dialogue generation problem, more interesting and challenging tasks relying on external knowledge have been explored to improve the anthropomorphic characteristic of dialogue systems.
%   have been explored to incorporate external knowledge into basic frameworks to enhance dialogue understanding and to improve anthropomorphic characteristic, which is emerging as a new fashion in the research of dialogue systems.
  A line of work introduced personalized information into dialogue generation to help deliver better dialogue response such as emotion~\cite{DBLP:conf/emnlp/LiS18,DBLP:conf/aaai/ZhouHZZL18,DBLP:conf/acl/SongZLXH19} and persona~\cite{DBLP:conf/acl/KielaWZDUS18,DBLP:conf/aaai/ZhengZHM20}.
  In addition, to further enhance and enrich the response generation, researchers have studied 
%   investigating 
  grounding dialogue generation on knowledge graphs~\cite{DBLP:conf/ijcai/ZhouYHZXZ18,DBLP:conf/acl/MoonSKS19} or unstructured documents~\cite{DBLP:conf/iclr/DinanRSFAW19,DBLP:conf/acl/KielaWZDUS18,DBLP:conf/emnlp/ZhouPB18,DBLP:journals/corr/abs-2010-10150,DBLP:journals/corr/abs-2012-11937}.

  \vspace{-1mm}
  \paragraph{Question Generation.}
  This task aims at generating a question from a given passage~\cite{DBLP:conf/acl/DuSC17} in an answer-aware or answer-unaware manner.
%   Another line of work focuses on answer-aware question generation by providing additional answer information.
  In this paper, we work on the answer-unaware setting, encouraging diversity of generated questions.
%   Given an answer and its associated context, it is possible to raise multiple questions with different focuses on the context and various means of expression.
  Researchers have explored statistical keyword extraction techniques to select salient words from input documents, and then incorporated the extracted keywords into question generation~\cite{DBLP:conf/emnlp/ChoSH19,DBLP:conf/emnlp/WangRZQTW20}
  Recent work has applied reinforcement learning to natural question generation~\cite{DBLP:conf/iclr/0022WZ20}.

  Different from previous text generation models that either incorporate unstructured Wikipedia knowledge or enhance the generation with exemplar cases, to the best of our knowledge, this paper makes the first attempt to retrieve and exploit both the task-specific and world knowledge for text generation in a unified framework. Our knowledge retrieval process is conducted through dense representations which can help to capture deep and latent semantics. 
\section{Method Formulation}
  The task of text generation is to output an appropriate target text given a source text as input.
  Given a dataset $\mathcal{D}$, an example is represented as $(s, t)$.
  Specifically, $s$ represents a source text and $t$ represents a target text. 
  A source text is used as a query to retrieve task-specific and world knowledge.
  Technically, the retrieved task-specific and world knowledge entries can be treated as two latent variables $z_1$ and $z_2$ respectively that are marginalized to get the Seq2Seq probability $p(t|s)$ via a top-\emph{m} approximation as
%   \begin{equation}
%     \begin{aligned}
%       p(t|s) &= \sum_{\substack{z_1 \in top-m(p_1(\cdot|s)), \\ z_2 \in top-m(p_2(\cdot|s)) }} p_1(z_1|s) \cdot p_2(z_2|s) \cdot p_\theta(t|s, z_1, z_2) \\
%              &= \sum_{\substack{z_1 \in top-m(p_1(\cdot|s)), \\ z_2 \in top-m(p_2(\cdot|s)) }} p_1(z_1|s) \cdot p_2(z_2|s) \cdot \prod_{t=1}^{|y|} p_\theta(t_k|s, z_1, z_2, t_{<k}),
%     \end{aligned}
%   \end{equation}
  \begin{equation}
    \begin{aligned}
      & p(t|s) = \sum_{\substack{z_1, z_2}} p_1(z_1|s) p_2(z_2|s) p_\theta(t|s, z_1, z_2) \\
             & = \sum_{\substack{z_1, z_2}} p_1(z_1|s) p_2(z_2|s) \prod_{t=1}^{|y|} p_\theta(t_j|s, z_1, z_2, t_{<j}),
    \end{aligned}
  \end{equation}
  where $z_1 \in$ top-\emph{m}$(p_1(\cdot|s))$, $z_2 \in$ top-\emph{m}$(p_2(\cdot|s))$, $t_j$ and $t_{<j}$ stand for the $j$-th token and the first $(j-1)$ tokens of a target text $t$ respectively, $|t|$ is the length of $t$, and the target text tokens are generated in an auto-regressive way.
  $p_1(\cdot|s)$ and $p_2(\cdot|s)$ are modeled with the retrieval probability that will be introduced in Eq.~(\ref{equ-score}).
  
\section{\MODELNAME{} Model}
  
  \begin{figure*}[t]
    \centering
    \includegraphics[width=15cm]{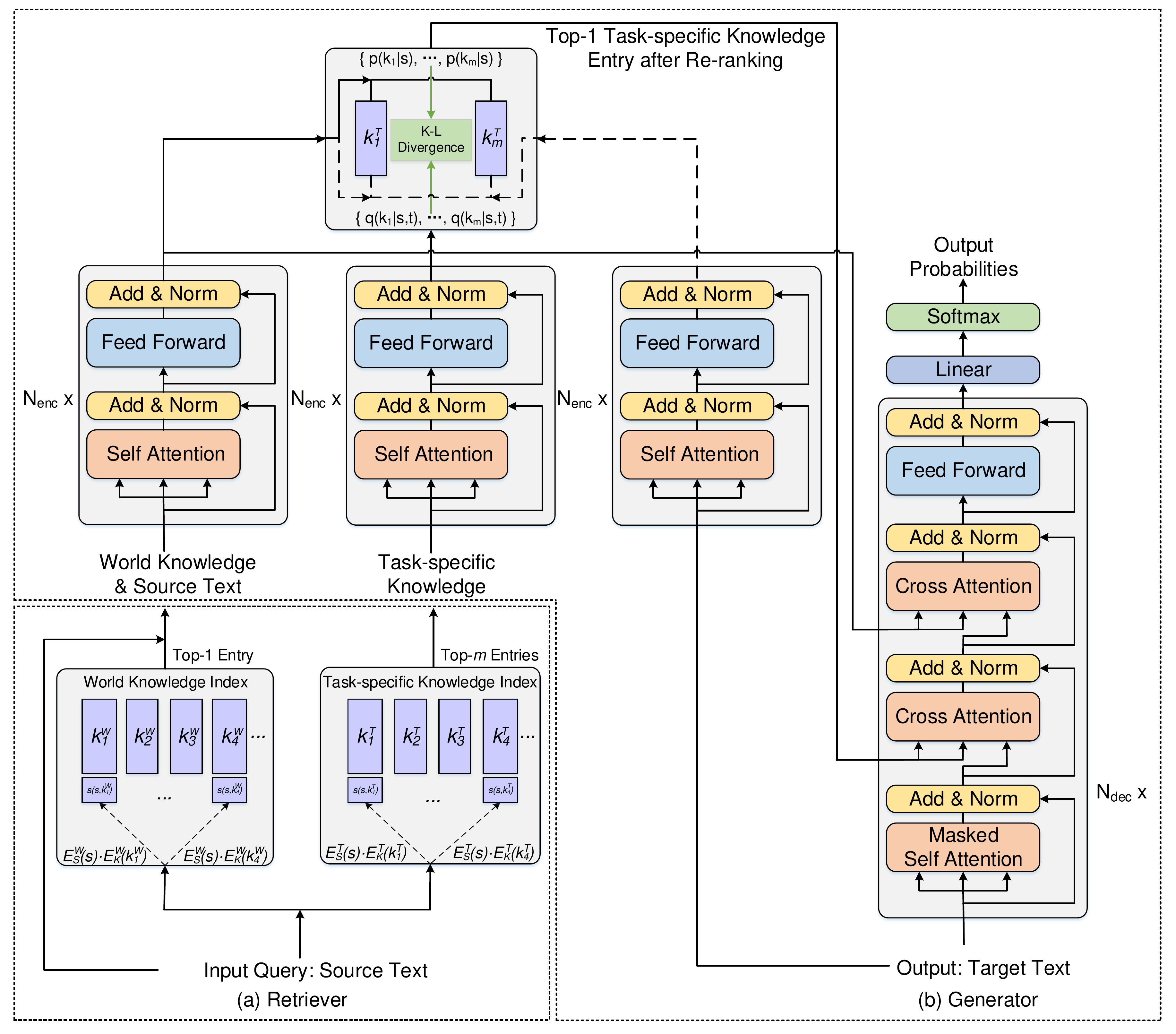}
    \vspace{-3mm}
    \caption{The overview architecture of our proposed \MODELNAME{} model which consists of (a) a retriever and (b) a generator. Here, $E^\alpha_\beta(\cdot)$ denotes the dense representation of an input sequence, where $\alpha \in$ \{world knowledge (\emph{W}), task-specific knowledge (\emph{T})\} and $\beta \in$ \{source text (\emph{S}), knowledge (\emph{K})\}.}
    % \vspace{-4mm}
    \label{fig-model}
  \end{figure*}

  Figure~\ref{fig-model} shows the overview architecture of \MODELNAME{} which consists of a \emph{retriever} and a \emph{generator}. 
  The retriever uses the input source text as a query to retrieve 
%   the task-specific and world knowledge
  the world knowledge and task-specific knowledge, the former of which is concatenated with the source text as additional background knowledge and the latter is fed into the decoder as exemplary information to guide the target text decoding. 
  Details about each component are provided in the following subsections.

  \subsection{Knowledge Retriever}  \label{sec-retriever}
  As shown in Figure~\ref{fig-model}(a), given a collection of a large number of knowledge entries ($k^\alpha_i$), the goal of the retriever is to index all knowledge entries in a low-dimensional and continuous space, so that it can retrieve efficiently the top-\emph{m} knowledge entries relevant to the input source text. Here, $\alpha \in$ \{world knowledge (\emph{W}), task-specific knowledge (\emph{T})\}.
  Inspired by the dense passage retrieval (DPR) \cite{DBLP:conf/emnlp/KarpukhinOMLWEC20}, we adopt a bi-encoder architecture to derive the dense representations of the source text and each knowledge entry. 
  Specifically, two independent pre-trained language models (i.e., BERT \cite{DBLP:conf/naacl/DevlinCLT19}), $E^{\alpha}_S(\cdot)$ and $E^{\alpha}_K(\cdot)$ are employed as the encoders for the source text and the knowledge entry respectively. 
  Furthermore, the representation of the \texttt{[CLS]} token is output as the dense representation. 
  At retrieval-time, the retriever first maps the input source text to a vector, and then retrieves knowledge entries of which vectors are the closest to the source text vector. 
  The similarity $s(s, k^{\alpha}_i)$ between the source text $s$ and each knowledge entry $k^{\alpha}_i$ is defined using the dot product of their vectors as
    \begin{align}
    \label{equ-score}
    s(s, k^{\alpha}_i) = E^{\alpha}_S(s)^{\top} \cdot E^{\alpha}_K(k^{\alpha}_i), i \in \{1, 2, ... \}.
    \end{align}
  Due to the significant difference between the two types of knowledge, we employ two independent retrievers for these two knowledge indexes.

    \paragraph{World Knowledge Retriever}
    % 开放式知识覆盖广，对于开放式对话能够提供很好的背景知识并且引导生成informative的回复。本文中，收到open-domain QA任务的启发，我们将这类知识具体为来自于wiki的文档。因此，我们采用oepn-domain QA任务中提供的知识集合作为开放式文档知识的index源，并且我们采用DPR模型作为我们的知识检索模型，由于他在各种知识检索任务上取得了不错的效果。
    World knowledge usually covers a wide variety of domains and has been proven effective in improving informativeness of the generated texts through providing the relevant background knowledge in open-domain text generation~\cite{DBLP:conf/iclr/DinanRSFAW19,DBLP:conf/iclr/ZhaoWTX0020}.
    % and its corpus size can easily range from millions of documents (e.g., Wikipedia) to billions (e.g., the Web).
    Motivated by the success of open-domain question answering (QA)~\cite{DBLP:conf/icml/GuuLTPC20,DBLP:conf/emnlp/KarpukhinOMLWEC20,DBLP:conf/acl/LeeCT19}, %and retrieval-augmented question generation~\cite{lewis2020retrieval}, 
    we assume the open-world knowledge as documents from the Wikipedia dump.
    % In this paper, a Wikipedia dump is employed as the world knowledge source, which has shown usefulness on question answering tasks~\cite{DBLP:conf/icml/GuuLTPC20,DBLP:conf/emnlp/KarpukhinOMLWEC20,DBLP:conf/acl/LeeCT19} and question generation~\cite{lewis2020retrieval}. 
    Specifically, we adopt the Wikipedia dump provided in open-domain QA tasks as our open-world knowledge which is composed of over $21$ millions of passages segmented from the Wikipedia pages.
    The goal of this retriever is to retrieve a small number of documents relevant to the given source text. 
    % Given the matched (positive) context-document pairs and the mismatched (negative) ones that are randomly sampled from the whole document corpus, the training of the context and the document encoders follows the same steps as those used for training the task-specific knowledge retriever. 
    Meanwhile, we use the DPR model which is a pre-trained bi-encoder released by \citet{DBLP:conf/emnlp/KarpukhinOMLWEC20} as the world knowledge retriever in our paper, since it has achieved great performance on various knowledge-intensive tasks.\footnote{https://github.com/facebookresearch/DPR} 
    The retrieved top-1 Wikipedia document ($k^W$) is employed for augmenting source text which will be described in Section~\ref{sec-generator}.
    % \ch{We retrieve top-1 wiki knowledge ($k^W$) for augmenting source text which will be described in Section \ref{sec-generator}.}
    % In our experiments, the pre-trained bi-encoder released by \citet{DBLP:conf/emnlp/KarpukhinOMLWEC20} is used to initialize our world knowledge retriever to build the index.\footnote{https://github.com/facebookresearch/DPR}

    \paragraph{Task-specific Knowledge Retriever}
    In addition to the world knowledge, it would also be desirable to obtain the relevant task-specific knowledge to guide the text generation process, since open-domain texts are often grounded by more than one single type of knowledge perception.
    % In our paper, we specify the task-specific knowledge as a bound of response candidates similar to the pre-built index in the retrieval-based dialogue. 
    % The retriever selects related response candidates that can act as exemplary information for guiding the response decoding.
    These related task-specific examples from a pre-built index can also bring additional information associated with the given source messages and provide exemplary information for guiding the target text decoding.

    % To obtain the expert knowledge for a specific task, we employ the data that shares the similar distribution to the training dataset to construct a task-specific knowledge index. 

    % Particularly, 3M context-response pairs are randomly selected for constructing the index corpus and the remaining 384k pairs are used for training the generator. 
    % Thus, there is no overlap between the retriever index and the generator training.
    
    % the training set in  that is composed of 3.384M context-response pairs is adopted.
    % Here, the retrieved related responses from the pre-built index serve as the knowledge providing the exemplary information of what they have said or responded to before.
    
    Formally, given a training example represented as $(s, t^{+}, t^{-}_1, ..., t^{-}_n)$, where each instance contains one source text $s$ and one matched (positive) target text $t^{+}$, along with $n$ mismatched (negative) distractors $t^{-}_i$ that are randomly sampled from the whole corpus, we can define the training objective function of the task-specific knowledge retriever as 
    % \begin{align}
    %   L(s, t^{+}, t^{-}_1, ..., t^{-}_n) = -\log \frac {e^{s(s, t^{+})}} {e^{s(s, t^{+})} + \sum_{j=1}^{n} e^{s(s, t^{-}_j)} }.
    % \end{align}
    \begin{equation}
    \begin{aligned}
        &L(s, t^{+}, t^{-}_1, ..., t^{-}_n) \\
      = &-\log \frac {e^{s(s, t^{+})}} {e^{s(s, t^{+})} + \sum_{i=1}^{n} e^{s(s, t^{-}_i)} }.
    \end{aligned}
    \end{equation}
    % \ch{At testing time, the model retrieves top-m knowledge items ($k^T$) with the highest probability calculated by Eq.~(\ref{equ-score})}
    At testing time, the model retrieves the top-\emph{m} knowledge entries ($k^T$) with the highest similarities calculated by Eq.~(\ref{equ-score}).
    
  \subsection{Generator}  \label{sec-generator}
    It is based on the pre-trained Transformer-based encoder-decoder architecture, BART~\cite{DBLP:conf/nips/VaswaniSPUJGKP17}.
    To incorporate both types of knowledge during the source text encoding and the target text decoding stages respectively, we make several modifications as follows. 
    % and is initialized with BART (base) \cite{DBLP:conf/acl/LewisLGGMLSZ20} which has shown great performance on various text generation tasks. 
    
    \paragraph{Augmented Source Text Encoder}
    In order to incorporate the world knowledge into the source text encoding stage, we concatenate the source text with the retrieved world knowledge entry.
    Formally, the input sequence is organized as $\{\texttt{[BOS]}, k^{W}_1, ..., k^{W}_{l_{k^{W}}} \texttt{[EOS]}, s_1, ..., s_{l_s}, \texttt{[EOS]}\}$, where \texttt{[BOS]} and \texttt{[EOS]} denote \emph{begin-of-sentence} and \emph{end-of-sentence}, $k^{W}_1, ..., k^{W}_{l_{k^{W}}}$ and $s_1, ..., s_{l_s}$ denote the knowledge and source text tokens, and $l_{k^{W}}$ and $l_s$ denote the token numbers of knowledge and source text respectively.
    Then the input sequence is fed into the stacked attention layers~\cite{DBLP:conf/nips/VaswaniSPUJGKP17,DBLP:conf/acl/LewisLGGMLSZ20} by employing itself as \emph{query}, \emph{key} and \emph{value} as
    \begin{align}
      \textbf{S}^{l+1} = \textsc{AttenLayer}(\textbf{S}^l), 
    \end{align}
    where $l \in \{0, ..., L-1\}$ and each $\textsc{AttenLayer}$ includes operations of a self-attention layer and a feed forward layer, both of which are followed by a residual connection and a layer normalization.
    $\textbf{S}^l \in \mathbb{R}^{(l_{k^{W}} + l_s + 3) \times d}$ denotes the representation of the concatenated source text and world knowledge at the \emph{l}-th encoder layer, and $d$ denotes the dimension of the embedding vector. 
    The outputs of each encoder layer are utilized as the inputs of the next encoder layer. 
    In each layer of encoding, the world knowledge serves as additional background and fully interacts with the source text to incorporate the relevant information into their representations through multi-head attention operations.
    After stacked layers of encoding, it can help to better understand the source text and return the contextualized representations, which will be further used during the decoding stage.

    \paragraph{Task-specific Knowledge Encoder}
    Different from the BERT-based encoding in Section~\ref{sec-retriever} for retrieval, another encoder that is a component of the generator, is designed to encode the task-specific knowledge to derive its contextualized representations for generation. 
    % \ch{Since the previous DPR model encodes the text only for computing semantic similarity and the basic framework used by the DPR model is BERT instead of BART for our generation, we redesigned another encoder to encode the task-specific knowledge to derive its contextualized representations.}
    % Another encoder is designed to encode the task-specific knowledge to derive its contextualized representations. 
    % Formally, the retrieved top-\emph{m} task-specific knowledge entries are first concatenated and organized as $\{\texttt{[BOS]}, k_{11}, ..., k_{1l_1} \texttt{[EOS]}, k_{21}, ..., k_{2l_2}, \texttt{[EOS]}, ...\}$.
    Formally, each of the retrieved top-\emph{m} task-specific knowledge entries is organized as $\{\texttt{[BOS]}, k^T_{i,1}, ..., k^T_{i,l^T_i}, \texttt{[EOS]}\}, i \in \{1, ..., m\}$.\footnote{We did study concatenating the source text with the task-specific knowledge as well, but no further improvement can be achieved.}
    Then the input sequence is fed into another encoder that does not share parameters with the augmented source text encoder. 
    Finally, we denote $\textbf{K}^{T, l}_i$ as the representation of the $i$-th task-specific knowledge at the \emph{l}-th encoder layer.

    \paragraph{Task-specific Knowledge Re-ranking}
    % Inspired by \citet{DBLP:conf/ijcai/LianXWPW19}, the target text is utilized for knowledge re-ranking to get more accurate selection results.
    Since the target text cannot be foreseen at testing time,
    a latent variable model~\cite{DBLP:conf/acl/ZhaoZE17,DBLP:conf/ijcai/LianXWPW19,kim2020sequential} is introduced to select the target text by treating it as the posterior information. 
    However, it is inefficient to calculate the prior and posterior probabilities in a large-scale dataset.
    Therefore, a task-specific knowledge re-ranking is designed for the top-\emph{m} knowledge entries output by the knowledge retriever.
    % \ch{Since the target text cannot be foreseen at testing time, to utilize the information of the target text, VAE~\cite{DBLP:conf/acl/ZhaoZE17} is introduced by treating the target text as the a posterior information. However, the computation of VAE in large-scale candidates is inefficient and the prior probabilities are difficult to fit the posterior. Therefore, the task-specific knowledge re-ranking is conducted based on the top-m knowledge selected in the first part. In general,}
    In general, to further calculate the similarity between each task-specific knowledge and the target text at a fine granularity, the target text is used for re-ranking the set of retrieved task-specific knowledge entries.
    The target text is encoded to acquire its representation, and then combined with the representation of the augmented source text to get the posterior representation, followed by a linear transformation as
    \begin{equation}
    %   \boldsymbol{C}(s, t) = \boldsymbol{W_c}[\boldsymbol{S}^{L-1}_{\texttt{[BOS]}}; \boldsymbol{T'}^{L-1}_{\texttt{[BOS]}}] + \boldsymbol{b_c},
      \textbf{c}(s, t) = \textbf{W}_c [\textbf{s}^{L}_{\texttt{[BOS]}}; \textbf{t}^{L'}_{\texttt{[BOS]}}] + \textbf{b}_c,
    \end{equation}
    where $\textbf{s}^{L}_{\texttt{[BOS]}}$ and $\textbf{t}^{L'}_{\texttt{[BOS]}}$ denote the outputs of the augmented source encoder and the target encoder corresponding to the \texttt{[BOS]} token, $\textbf{W}_c$ and $\textbf{b}_c$ are parameters updated during training.
    % the same architecture as task-specific knowledge encoder, 
    The similarity between this representation and the representation of each task-specific knowledge entry is calculated to obtain the probability distribution of re-ranking,
    \begin{equation}
      q_{\phi}(k^T_i|s, t) = \softmax(\textbf{c}(s, t) \cdot \textbf{k}^{T, L}_{i, \texttt{[BOS]}}), 
    \end{equation}
    for $i \in \{1, ..., m \}$.
    In order to accommodate the situation where the target text is not available when testing, the prior probability is calculated as
    \begin{equation}
      p_{\theta}(k^T_i|s) = \softmax( \textbf{s}^{L}_{\texttt{[BOS]}} \cdot \textbf{k}^{T, L}_{i, \texttt{[BOS]}}),
    \end{equation}
    for $i \in \{1, ..., m \}$.
    Finally, two probability distributions of $q_{\phi}(\boldsymbol{k}^T|s, t)$ and $p_{\theta}(\boldsymbol{k}^T|s)$ are approximated in a way optimizing KL divergence as
    \begin{equation} \label{equ: KL}
    \begin{aligned}
    %   \mathcal{L}_{kl} = \mathbb{E}_{k\sim{q_{\phi}(\boldsymbol{k}|s, t)}} \log{\frac{q_{\phi}(\boldsymbol{k}|s, t)}{p_{\theta}(\boldsymbol{k}|s)}}.
    \mathcal{L}_{kl} = \mathbb{E}_{q_{\phi}(\boldsymbol{k}^T|s, t)} \log{\frac{q_{\phi}(\boldsymbol{k}^T|s, t)}{p_{\theta}(\boldsymbol{k}^T|s)}}.
    \end{aligned}
    \end{equation}
    The bag-of-words (BOW) loss~\cite{DBLP:conf/acl/ZhaoZE17} is introduced to facilitate the training process as
    \begin{equation}
      \mathcal{L}_{bow} = - \mathbb{E}_{k^T\sim{q_{\phi}(\boldsymbol{k}^T|s, t)}} \sum^{l_t}_{j=1} \log{p(t_{j}|k^T)},
    \end{equation}
    where $p(t_j|k^T)$ denotes the estimated probability of word $t_j$ calculated by
    \begin{equation}
      p(\boldsymbol{\cdot}|k^T) = \softmax(\textbf{W}_{\texttt{bow}} \textbf{k}^{T, L}_{\texttt{[BOS]}} + \textbf{b}_{\texttt{bow}}),
    \end{equation}
    where $\textbf{k}^{T, L}_{\texttt{[BOS]}}$ denote the outputs of the knowledge encoder corresponding to the \texttt{[BOS]} token of the selected knowledge, $\textbf{W}_{\texttt{bow}}$ and $\textbf{b}_{\texttt{bow}}$ are parameters updated during training.

    \paragraph{Decoder}
    % The decoder is also composed of a stack of identical layers.
    % \footnote{We did study swapping the order of attending to outputs of two encoders, and preliminary results showed that first attending to the output of the task-specific knowledge encoder achieved a slightly better performance.} 
    In order to inject all the encoded information of the source text, the world knowledge and the task-specific knowledge to guide the target text decoding, two additional sub-layers are inserted into each decoder layer, which perform cross-attention over the output of the last layer of the two encoders. 
    Particularly, after a sub-layer of masked self-attention where each token cannot attend to future tokens to avoid information leakage, the target text first attends to the output of the task-specific knowledge encoder and then attends to the output of the augmented source text encoder.
    Mathematically, we have
    \begin{equation}
    \begin{aligned}
      \bar{\textbf{T}}^l   &= \textsc{LN} \Big(\textbf{T}^l + \textsc{SelfAtten}(\textbf{T}^l) \Big),  \\
      \tilde{\textbf{T}}^l &= \textsc{LN} \Big(\bar{\textbf{T}}^l + \textsc{CrossAtten} (\bar{\textbf{T}}^l, \textbf{K}^{T, L})\Big),  \\
      \hat{\textbf{T}}^l   &= \textsc{LN} \Big( \tilde{\textbf{T}}^l+ \textsc{CrossAtten} (\tilde{\textbf{T}}^l, \textbf{S}^L)\Big), \\
      \textbf{T}^{l+1}     &= \textsc{LN} \Big(\hat{\textbf{T}}^l + \textsc{FeedForward} (\hat{\textbf{T}}^l)\Big),
    \end{aligned}
    \end{equation}
    where $l \in \{0, \cdots, L-1\}$, $\textsc{LN}$ denotes the operation of layer normalization, $\textbf{T}^l$ denotes the representation of the target text at the \emph{l}-th decoder layer, $\bar{\textbf{T}}^l$, $\tilde{\textbf{T}}^l$ and $\hat{\textbf{T}}^l$ are intermediate representations after each operation. 
    In this way, the model can first learn \emph{how to generate} and consider the retrieved task-specific knowledge as exemplary information.
    The model can further learn \emph{what to say} according to the retrieved world knowledge that is used to augment the source text and enrich the exemplary information.

  \subsection{Learning}
    Given the representation of each target text token at the last decoder layer $\textbf{T}^{L} = \{\textbf{t}_{j}\}_{j=1}^{l_t}$ where $\textbf{t}_{j} \in \mathbb{R}^{d}$, the probability distribution over the whole vocabulary of each target text token $\textbf{p}_{t_{j}}$ can be calculated via a non-linear transformation.
    % \begin{align}
    %   \textbf{p}_{r_{t}} = \softmax ( \textbf{r}_{t} \cdot \textbf{W}_{vocab} + \textbf{b}_{vocab} ),
    % \end{align}
    % where $ \textbf{W}_{vocab} \in \mathbb{R}^{d \times V} $ is the token embedding table, $V$ denotes the vocabulary size, and $\textbf{b}_{vocab} \in \mathbb{R}^{V} $ is a bias vector.
    The learning objective of this task is to minimize the negative log-likelihood loss as
    % \begin{equation}
    %   \mathcal{L}_{lm} = - \frac{1}{l_t} \sum_{k=1}^{l_t} log~p_{t_{k}},
    % \end{equation}
    \begin{equation}
      \mathcal{L}_{gen} = - \mathbb{E}_{k^T\sim{q_{\phi}(\boldsymbol{k}^T|s, t)}} \sum^{l_t}_{j=1} \log{p(t_{j}|s, t_{<j}, k^T)}.
    \end{equation}
    % where $p_{t_{j}}$ is the element in \textbf{p}$_{t_{j}}$ corresponding to the original token. 
    Finally, the parameters of our model are optimized by performing multi-task learning by minimizing the sum of all loss functions as
    \begin{equation} \label{equ: Loss}
      \mathcal{L}_{total} =  \mathcal{L}_{gen} + \mathcal{L}_{kl} + \mathcal{L}_{bow}.
    \end{equation}

\section{Experiments}

  We evaluated the proposed method on the tasks of dialogue generation and question generation.

    %%%%%%%%%%%%%%%%%%%%%%%%%%%%%%%%%%%%%%%%%%%%%%%%%%%%%%%%%%%%%%%%%%%%%%%%%%%%%%%%%%%%%%%%%%%

    % \begin{table*}[t] 
    % %   \vspace{-2mm}
    %   \centering
    %   \resizebox{0.95\linewidth}{!}{
    %   \begin{tabular}{lcccccc}
    %   \toprule
    %     \backslashbox{Datasets}{Statistics}       & Task & Index type & Index size & Train & Valid & Test  \\
    %   \midrule 
    %     Reddit \cite{DBLP:conf/ijcai/ZhouYHZXZ18} & Dialogue generation & Context-response pair & 3M  & 38.4k & 10k  & 20k \\
    %     WoW \cite{DBLP:conf/iclr/DinanRSFAW19}    & Dialogue generation & Context-response pair & 34k & 40k   & 3.9k & 3.9k \\
    %     SQuAD \cite{DBLP:conf/acl/DuSC17}         & Question generation & Sentence-question pair & 45k & 25.5k & 10.5k & 11.9k \\
    %   \bottomrule
    %   \end{tabular}
    %   }
    %   \caption{Statistics of the task-specific knowledge indexes and the datasets used in our experiments.}
    % %   \vspace{-4mm}
    %   \label{tab-data}
    % \end{table*}

    %%%%%%%%%%%%%%%%%%%%%%%%%%%%%%%%%%%%%%%%%%%%%%%%%%%%%%%%%%%%%%%%%%%%%%%%%%%%%%%%%%%%%%%%%%%

    \begin{table*}[t] 
    %   \vspace{-2mm}
      \centering
      \resizebox{0.9\linewidth}{!}{
      \begin{tabular}{lccccccc}
      \toprule
        \backslashbox{Models}{Metrics}                       & BLEU-1 & BLEU-2 & METEOR & ROUGE$_L$ & Average & Greedy & Extrema \\
      \midrule 
        RNN \cite{DBLP:conf/nips/SutskeverVL14}              & 7.36 & 2.94 & 7.28 & 10.03 & 0.6591 & 2.0585 & 0.3331  \\
        CVAE \cite{DBLP:conf/acl/ZhaoZE17}                   & 7.45 & 2.85 & 7.34 & 9.68  & 0.6642 & 2.0853 & 0.3357  \\
        Transformer \cite{DBLP:conf/nips/VaswaniSPUJGKP17}   & 7.97 & 3.14 & 7.92 & 10.51 & 0.6693 & 2.0703 & 0.3334  \\
        GPT-2 \cite{radford2019language}                     & 8.43	& 3.04 & 8.33 & 10.65 & 0.6484 & 2.0601 & 0.3303  \\
        DialoGPT \cite{DBLP:conf/acl/ZhangSGCBGGLD20}        & 7.58 & 3.02 & 7.85 & 10.82 & 0.5976 & 2.0774 & 0.3185 \\ %??
        BART \cite{DBLP:conf/acl/LewisLGGMLSZ20}             & 9.24 & 3.38 & 9.03 & 10.93 & 0.6611 & 2.0986 & 0.3355  \\        
      \midrule 
        % \MODELNAME{}                                       & \textbf{9.59} & \textbf{3.47} & \textbf{9.40} & \textbf{11.12} & \textbf{0.6728} & \textbf{2.1114} & \textbf{0.3397} \\
        \MODELNAME{}                                         & \textbf{9.71} & \textbf{3.63} & \textbf{9.53} & \textbf{11.36} & 0.6522 & \textbf{2.1683} & 0.3362 \\
      \midrule 
        % \MODELNAME{} w/o. WK                                       & 9.32 & 3.38 & 9.14 & 11.04 & 0.6596 & 2.0833 & 0.3336 \\
        \MODELNAME{} w/o. WK                                 & 9.52 & 3.58 & 9.44 & 11.32 & 0.6490 & 2.1647 & 0.3361 \\
        \MODELNAME{} w/o. TK                                 & 9.35 & 3.39 & 9.06 & 11.02 & 0.6644 & 2.0968 & 0.3371 \\
      \bottomrule
      \end{tabular}
      }
      \caption{Performance of our method and previous methods on the test set of Reddit dataset for dialogue generation \cite{DBLP:conf/ijcai/ZhouYHZXZ18} in terms of the automated evaluation metrics. Numbers in bold denote that the improvement over the best performing baseline is statistically significant (t-test with \emph{p}-value $<$ 0.05). WK and TK denote world knowledge and task-specific knowledge respectively.}
    %   \vspace{-4mm}
      \label{tab-result-reddit-auto}
    \end{table*}

    %%%%%%%%%%%%%%%%%%%%%%%%%%%%%%%%%%%%%%%%%%%%%%%%%%%%%%%%%%%%%%%%%%%%%%%%%%%%%%%%%%%%%%%%%%%

    \begin{table*}[t] 
    %   \vspace{-2mm}
      \centering
      \resizebox{0.9\linewidth}{!}{
      \begin{tabular}{lcccccc}
      \toprule
        \backslashbox{Models}{Metrics}                       & BLEU-1 & BLEU-2 & BLEU-3 & BLEU-4 & METEOR & ROUGE$_L$  \\
      \midrule 
        Vanilla seq2seq \cite{DBLP:conf/nips/SutskeverVL14}  & 31.34 & 13.79 & 7.36 & 4.26 & 9.88 & 29.75  \\
        H\&S \cite{DBLP:conf/acl/DuSC17}                     & 38.50 & 22.80 & 15.52 & 11.18 & 15.95 & 30.98  \\
        NQG \cite{DBLP:conf/acl/DuSC17}                      & 43.09 & 25.96 & 17.50 & 12.28 & 16.62 & 39.75  \\
        BART \cite{DBLP:conf/acl/LewisLGGMLSZ20}             & 45.16 & 29.45 & 21.33 & 16.09 & 19.70 & 43.44  \\        
      \midrule 
        \MODELNAME{}                                         & \textbf{46.57} & \textbf{30.64} & \textbf{22.28} & \textbf{16.75} & \textbf{20.37} & \textbf{43.63}  \\
      \midrule 
        \MODELNAME{} w/o. WK                                 & 46.25 & 30.29 & 21.94 & 16.49 & 20.10 & 43.43  \\
        \MODELNAME{} w/o. TK                                 & 45.63 & 30.02 & 21.88 & 16.56 & 19.79 & 43.61  \\
      \bottomrule
      \end{tabular}
      }
      \caption{Performance of our method and previous methods on the test set of SQuAD dataset for question generation \cite{DBLP:conf/acl/DuSC17} in terms of the automated evaluation metrics.}
    %   \vspace{-4mm}
      \label{tab-result-squad-auto}
    \end{table*}

    %%%%%%%%%%%%%%%%%%%%%%%%%%%%%%%%%%%%%%%%%%%%%%%%%%%%%%%%%%%%%%%%%%%%%%%%%%%%%%%%%%%%%%%%%%%

    \begin{table}[t]%[!hbt]
      \centering
      \resizebox{0.98\linewidth}{!}{
      \begin{tabular}{lcccc}
      \toprule
        % \backslashbox{Models}{Aspects}                       & Relevance & Fluency & Informativeness & Kappa  \\
        \backslashbox{Models}{Aspects}                       & Rel. & Flu. & Inform. & Kappa  \\
      \midrule 
        Human                                                &  1.40     &  1.64   &    1.47         & 0.62   \\
      \midrule
        % Transformer \cite{DBLP:conf/nips/VaswaniSPUJGKP17}   &  0.86     &  1.07   &    0.71         & 0.42   \\
        % GPT-2 \cite{radford2019language}                     &  1.09     &  1.20   &    0.84         & 0.43   \\
        % BART \cite{DBLP:conf/acl/LewisLGGMLSZ20}             &  1.36     &  1.48   &    1.14         & 0.47   \\
        % \MODELNAME{}                                         &  1.44     &  1.51   &    1.23         & 0.46   \\
        Transformer  &  0.86     &  1.07   &    0.71         & 0.42   \\
        GPT-2        &  1.09     &  1.20   &    0.84         & 0.43   \\
        BART         &  1.36     &  1.48   &    1.14         & 0.47   \\
        \MODELNAME{} &  1.44     &  1.51   &    1.23         & 0.46   \\
      \bottomrule
      \end{tabular}
      }
      \caption{
      Human evaluation results of \MODELNAME{} on a randomly sampled test set of the Reddit dataset.
      Here, Rel., Flu., and Inform. indicates relevance, fluency, and informativeness respectively.
      } % and selected systems
      \vspace{-4mm}
      \label{tab-result-reddit-human}
    \end{table}

    %%%%%%%%%%%%%%%%%%%%%%%%%%%%%%%%%%%%%%%%%%%%%%%%%%%%%%%%%%%%%%%%%%%%%%%%%%%%%%%%%%%%%%%%%%%

  \subsection{Knowledge and Datasets}
  
    \paragraph{World Knowledge Index.}
    For the world knowledge, all tasks and datasets shared the same English Wikipedia dump from Dec. 20, 2018 provided by \citet{DBLP:conf/acl/LeeCT19}. 
    Each Wikipedia article was split into disjoint 100-word chunks to make a total of 21M documents.
    % Each passage was also prepended with the title of the Wikipedia article where the passage was from, along with an \texttt{[SEP]} token. 
    Each passage was also prepended with its title, along with an \texttt{[SEP]} token. 

    \paragraph{Reddit Dataset for Dialogue Generation.}
    To construct the task-specific knowledge index for this dataset, the Reddit dialogue corpus collected by \citet{DBLP:conf/ijcai/ZhouYHZXZ18} was used. 
    $3$ millions responses were randomly sampled from the training set of the Reddit dataset. 
    After excluding the samples used for constructing the task-specific knowledge index, the remaining dataset composed of 38.4k/10k/20k context-response pairs in the training/validation/testing sets respectively, was employed to train a generator and to evaluate the performance of our framework. 
    Thus, there is no data overlap between that for the task-specific knowledge index and that for learning a generator.
    % \footnote{http://coai.cs.tsinghua.edu.cn/file/commonsense\_conversation\_dataset.tar.gz}
    
    \paragraph{SQuAD Dataset for Question Generation.}
    Similarly, 45k randomly selected sentence-question pairs from the training set of the SQuAD Dataset processed by \citet{DBLP:conf/acl/DuSC17} were used to construct the task-specific knowledge index for this dataset.
    Also, the remaining dataset composed of 25.5k/10.5k/11.9k sentence-question pairs in the training/validation/testing sets respectively, was employed to train the generator. 
    
    % \paragraph{Task-specific Knowledge Index.}
    % To construct the task-specific knowledge index for each dataset, the context-response or sentence-question pairs were randomly sampled from the training set of each dataset.
    % After excluding the samples used for constructing the task-specific knowledge index, the remaining dataset was employed to train the generator and to evaluate the performance of our framework. 
    % Thus, there is no overlap between data for the task-specific knowledge index and data for the learning of generator.
    % The statistics of the task-specific knowledge indexes and the datasets used in our experiments were shown in Table~\ref{tab-data}.
    % Readers can refer to Appendix~\ref{sec-data} for more details.

  \subsection{Baseline Models}
    % The following models were selected as the baseline models:
    % RNN \cite{DBLP:conf/nips/SutskeverVL14},
    % CVAE \cite{DBLP:conf/acl/ZhaoZE17},
    % Transformer \cite{DBLP:conf/nips/VaswaniSPUJGKP17},
    % GPT-2 \cite{radford2019language}, 
    % DialoGPT \cite{DBLP:conf/acl/ZhangSGCBGGLD20}, and
    % BART \cite{DBLP:conf/acl/LewisLGGMLSZ20}.
    % Readers can refer to \ref{sec-baseline} for baseline implementation details.
    The following models were selected as the baseline models:
    \textbf{(1) RNN} \cite{DBLP:conf/nips/SutskeverVL14} is a LSTM-based sequence-to-sequence model with attention mechanism.
    \textbf{(2) CVAE} \cite{DBLP:conf/acl/ZhaoZE17} uses latent variables to learn a distribution over potential conversation contexts based on conditional variational autoencoders.
    \textbf{(3) Transformer} \cite{DBLP:conf/nips/VaswaniSPUJGKP17} uses the self-attention mechanism to build the encoder and the decoder, which has shown better performance than RNN-based Seq2Seq models in many natural language processing tasks. 
    \textbf{(4) GPT-2} \cite{radford2019language} is a uni-directional pre-trained language model that has shown great performance on a lot of natural language generation tasks. Following its original concatenation operation, the context and the response were concatenated with a special \texttt{[SEP]} token as input for encoding.
    \textbf{(5) DialoGPT} \cite{DBLP:conf/acl/ZhangSGCBGGLD20} has the same architecture with GPT-2 but is trained with Reddit discussions Datasets.
    \textbf{(6) BART} \cite{DBLP:conf/acl/LewisLGGMLSZ20} is a denoising autoencoder using a standard Tranformer-based neural machine translation architecture for pre-training the sequence-to-sequence models. %which can be seen as generalizing BERT (due to the bidirectional encoder), GPT (with the left-to-right decoder), and other recent pre-training schemes. 
    BART is trained by corrupting text with an arbitrary noising function to reconstruct the original text.

  \subsection{Evaluation Metrics}
    To ensure all experimental results were comparable, the automated and human evaluation metrics popular used in previous work were adopted in this paper. 
    BLEU, METEOR, ROUGE$_L$ and three embedding-based metrics including Embedding Average, Greedy Matching and Extrema Score used in \citet{forgues2014bootstrapping} which can cover the weaknesses of BLEU were employed as the automated metrics. 
    Human evaluation was also conducted to measure the quality of the generated responses of models in terms of three independent aspects: 1) relevance (Rel.), 2) fluency (Flu.) and 3) informativeness (Inform.). 
    Each judge was asked to give three scores for a response, each of which was ranged from 0 to 2.

  \subsection{Training Details}
    % Readers refer to Appendix~\ref{sec-implement} for training details.
    Model parameters were initialized with pre-trained weights of \emph{bart-base} released by \citet{DBLP:conf/emnlp/WolfDSCDMCRLFDS20}. 
    The word embedding table was shared between the encoder and decoder.
    The AdamW method~\cite{DBLP:conf/iclr/LoshchilovH19} was employed for optimization. 
    The learning rate was initialized as $6.25e\text{-}5$ and was decayed linearly down to $0$. 
    The max gradient norm was clipped down to $1.0$.
    The batch size was set to $64$. %with $8$ gradient accumulation steps.
    The maximum length of the concatenation of open-domain knowledge and context was set to $128$. 
    The maximum length of the task-specific knowledge was set to $128$. 
    The number of task-specific knowledge entries was set to 3, achieving the best performance out of \{1, 2, 3, 4, 5\} on the validation set.
    The strategy of greedy search was performed for decoding. 
    The maximum length of response to generate was also set to $50$. 
    All experiments were run on a single A100 GPU. 
    The maximum number of epochs was set to 15. 
    The validation set was used to select the best model for testing. 
    All code was implemented in the PyTorch framework\footnote{https://pytorch.org/} and are published to help replicate our results.~\footnote{https://github.com/lxchtan/TEGTOK}
  
  \subsection{Evaluation Results}
    
    \paragraph{Automated Evaluation}
    Table~\ref{tab-result-reddit-auto} and Table~\ref{tab-result-squad-auto} present the evaluation results of our method and previous methods on the test sets of the Reddit dataset for dialogue generation and the SQuAD dataset for question generation respectively. 
    Each model ran four times with identical architectures and different random initializations, and the best out of them was reported. 
    The results show that our method outperformed all baseline models in terms of all metrics. 
    % Specifically, compared with GPT-2, our method outperformed it by margins of 1.16\% in terms of BLEU-1 and 1.07\% in terms of METEOR.
    % Compared with BART, our method outperformed it by margins of 0.35\% in terms of BLEU-1 and 0.37\% in terms of METEOR, illustrating the importance of incorporating the world knowledge and the task-specific knowledge. 
    Specifically, \MODELNAME{} outperformed GPT-2 by 1.28\% BLEU-1 and 1.20\% METEOR, outperformed DialoGPT by 2.13\% BLEU-1 and 1.68\% METEOR, and outperformed BART by 0.47\% BLEU-1 and 0.50\% METEOR on the Reddit dataset. 
    Meanwhile, \MODELNAME{} outperformed BART by 1.41\% BLEU-1 and 0.67\% METEOR on the SQuAD dataset, illustrating the effectiveness of incorporating 
    both two types of knowledge.
    % the task-specific and open-world knowledge. 
    
    To further verify the effectiveness of each component in our proposed methods, ablation tests were conducted as shown in the last two rows of Table~\ref{tab-result-reddit-auto} and Table~\ref{tab-result-squad-auto}. 
    First, the world knowledge was ablated and the results show that BLEU-1 and METEOR dropped down by 0.27\% and 0.26\% respectively on the Reddit dataset, along with 0.32\% and 0.27\% respectively on the SQuAD dataset, illustrating the effectiveness of retrieving world knowledge for text generation. 
    On the other hand, the task-specific knowledge was ablated and only the world knowledge can be attended to during the decoding stage. 
    The results show that BLEU-1 and METEOR dropped down by 0.24\% and 0.34\% respectively on the Reddit dataset, along with 0.94\% and 0.58\% respectively on the SQuAD dataset, illustrating the effectiveness of attending to task-specific knowledge during the decoding stage.

    \paragraph{Human Evaluation}
    Table~\ref{tab-result-reddit-human} presents the human evaluation results on a randomly sampled test set of the Reddit dataset. 
    100 samples were evaluated and the order of evaluation systems were shuffled. 
    Three judges were asked to score from 0 to 2 (2 for the best) for each human evaluation aspect and the average scores were reported. 
    The Fleiss’s kappa value~\cite{fleiss1971measuring} for each model was also reported, indicating the inter-judge moderate agreement during evaluation.
    % indicating that the judges reached moderate agreement. 
    In general, the results show that our method outperformed all baseline models, showing that it can generate more natural responses. 
    Particularly, compared with BART, our method achieves the greatest improvement in terms of informativeness, illustrating the effectiveness of incorporating the task-specific and world knowledge for improving informativeness of generated texts.

  \subsection{Case Study}  
    
    % \begin{figure}[t]
    %   \centering
    %   \includegraphics[width=8cm]{figures/trend_k2_number.pdf}
    %   \caption{Performance of our model under different numbers of task-specific knowledge entries on the test set of the Reddit dataset.}
    %   \label{fig-k2-number}
    % \end{figure}
    
    % \paragraph{The impact of numbers of task-specific knowledge entries.}
    % TODO
    % Figure~\ref{fig-k2-number} illustrates how the performance of our method changed with respect to different numbers of task-specific knowledge entries on the test set of the Reddit dataset. 
    % It can be seen that the performance was significantly improved as the number of task-specific knowledge entries increased at the beginning, which shows the effectiveness of incorporating the task-specific knowledge entries for generation. 
    % Then, the performance became stable and dropped. 
    % The reason might be that incorporating information that is not so strongly correlated will introduce noise.

    \begin{table*}[t]
    \centering
    \small
    \begin{tabular}{p{0.95\linewidth}}
    \toprule
    \multicolumn{1}{c}{\textbf{Case 1}} \\
    \midrule
      \textbf{Context}:         whatever happened to al qaeda?  \\ 
      \textbf{WK}:              Al-Qaeda operates as a network of Islamic extremists and Salafist jihadists. The organization has been designated as a \textcolor{blue}{terrorist} group by the United Nations Security Council, ... The Taliban provided a safe haven for Osama bin Laden and al-Qaeda officials, allowing them to plot major terrorist attacks such as the September 11 attacks (9/11). ... \\
    %   TK:            & (1) isis first iteration was al - qaeda \textcolor{red}{in iraq}. (2) is isil the sequel to isis?, (3) so ? back then they just called them al - qaeda \textcolor{red}{in iraq}, does n't change anything.  \\
      \textbf{TK}:             isis first iteration was al - qaeda \textcolor{red}{in iraq}.  \\
    \midrule
      \textbf{Transformer}:     i 'm not sure what you 're talking about , but i 'm not sure if you 're referring to what you 're talking about. \\ 
      \textbf{GPT-2}:           i think he was a member of the al qaeda branch. \\ 
      \textbf{DialoGPT}:       they're still around. \\
      \textbf{BART}:           i'm not sure. i'm sure the media is talking about the death of the leader of the country. \\ 
      \textbf{\MODELNAME{}}:    they're a \textcolor{blue}{terrorist} organization \textcolor{red}{in iraq} plot major attacks.  \\
    \toprule
        \multicolumn{1}{c}{\textbf{Case 2}} \\
    \midrule
      \textbf{Passage}:         in late  \textcolor{blue}{summer} he was invited by jane \textcolor{blue}{stirling} to visit scotland , where he stayed at calder house near edinburgh and at johnstone castle in renfrewshire , both owned by members of stirling 's family .  \\ 
      \textbf{WK}:              ... After this, in \textcolor{blue}{1860} \textcolor{blue}{Stirling} returned to Edinburgh - his address there was 4 Laverock Bank Road, Trinity, Edinburgh - which then became his permanent residence until ...  \\
      \textbf{TK}:              \textcolor{red}{where did} victoria and her family retreat to safety during a conflict \textcolor{red}{in} 1848?  \\
    \midrule
      \textbf{BART}:            where was johnstone castle? \\ 
      \textbf{\MODELNAME{}}:    \textcolor{red}{where did} \textcolor{blue}{stirling} stay \textcolor{red}{in} the  \textcolor{blue}{summer} of \textcolor{blue}{1860}? \\
    % \bottomrule
    \bottomrule
    \end{tabular}
    \caption{
    % Case 1 is the dialogue generation result of a test sample in the Reddit dataset.
    % Case 2 is the question generation result in the SQuAD dataset. 
    Generation results of two cases from the Reddit and SQuAD datasets respectively.
    % We kept original texts without manual corrections following objective facts. 
    We kept original texts without manual corrections.
    WK and TK denote world knowledge and task-specific knowledge respectively. Words in the same color are related.
    } %following objective facts.
    \label{tab-case-squad}
    \end{table*}    
    
    Case studies were conducted by randomly sampling an instance from the Reddit dataset in dialogue generation and an instance from the SQuAD dataset in question generation as shown in Table~\ref{tab-case-squad}. 
    % As the sample shown in Table 4, world knowledge provides information about entities for the generated text, while task-specific knowledge provides relevant syntactic information.
    Given the conversation context (or the passage of a question), it was used as a query to retrieve the task-specific and world knowledge in the upper block of a single case in Table~\ref{tab-case-squad}.
    For case 1, as we can see that, there was no text overlap between the second task-specific knowledge entry and the conversation context, but it can be retrieved through semantic relevance, which shows the effectiveness of using dense representations for knowledge retrieval.
    Since the given context is short and contains few informative words, it is difficult for models to generate informative responses without any external knowledge, such as the generic response generated by the Transformer model. 
    Furthermore, our generated response can capture the relevant and important information from the retrieved knowledge, such as ``\emph{terrorist}'' from the world knowledge and ``\emph{in iraq}'' from the task-specific knowledge, making the generated response more informative and illustrating the effectiveness of incorporating these two types of knowledge for dialogue generation. 
    For case 2, we can see that there was little text overlap between the world knowledge and the passage, but it could be retrieved through semantic relevance, showing the effectiveness of using dense representations for knowledge retrieval.
    Our generated text can capture the relevant and important information from the retrieved world knowledge, such as ``\emph{1860}'' and ``\emph{Stirling}'' from the world knowledge, making the generated text more informative. 
    Furthermore, since the given passage mainly focuses on narrative descriptions, it is difficult for models to generate exemplar texts without any external knowledge, such as the ``\emph{where did ... in}'' question template retrieved from the task-specific knowledge index.
    Again, these results illustrated the effectiveness of incorporating these two types of knowledge for question generation.

\section{Conclusion}
  In this paper, we study retrieving relevant external knowledge for enhancing text generation. 
  Two types of knowledge, i.e., task-specific and world knowledge, are retrieved using dense representations to ensure effectiveness and efficiency of knowledge selection, and are further incorporated into the input encoding and output decoding stages respectively, providing the supplementary information to guide text generation. 
  Experimental results on two tasks of dialogue generation and question generation show that our method achieves better performance than baseline models and can generate more informative texts.
  In the future, we will explore applying this framework to more text generation tasks and other modalities such as image caption, to further verify its effectiveness and generalization.

\section*{Acknowledgements}
  We thank anonymous reviewers for their valuable comments. 

% \clearpage
\bibliography{custom}
\bibliographystyle{acl_natbib}

% \clearpage
% \appendix
% \input{appendix}

\end{document}